\documentclass[letterpaper, 10 pt, conference]{ieeeconf}  %

\IEEEoverridecommandlockouts                              %

\usepackage{graphics} %
\usepackage{epsfig} %
\usepackage{fontawesome5}
\usepackage{booktabs}
\usepackage{multirow}
\usepackage{graphicx}
\usepackage{siunitx}
\usepackage{amsmath} %
\usepackage{amssymb}  %
\usepackage{cite}
\usepackage{wrapfig}
\usepackage{stfloats}
\usepackage{subcaption}
\usepackage{soul}
\setstcolor{red}

\makeatletter
\let\NAT@parse\undefined
\makeatother
\usepackage[hidelinks]{hyperref}

\title{\LARGE \bf
Agile Flight Emerges from Multi-Agent Competitive Racing 
}

\author{Vineet Pasumarti$^*$  {}  Lorenzo Bianchi$^*$ {} \thanks{$^*$ denotes equal contribution. LB is affiliated to the University of Rome Tor Vergata. VP and AL are with the University of Pennsylvania.}%
 {}  Antonio Loquercio} %
\usepackage{xcolor}
\usepackage{xspace}
\usepackage{bm}

\newcommand{\thor}{T}

\newcommand{\ego}{ego\xspace}
\newcommand{\adversary}{adversary\xspace}
\newcommand{\E}{\mathrm{e}}
\newcommand{\A}{\mathrm{a}}
\newcommand{\xe}{\bm{x}^\E}
\newcommand{\xa}{\bm{x}^\A}
\newcommand{\ue}{\bm{u}^\E}
\newcommand{\ua}{\bm{u}^\A}

\begin{document}

\maketitle
\thispagestyle{empty}
\pagestyle{empty}

\begin{abstract}

Through multi-agent competition and the sparse high-level objective of winning a race, we find that both agile flight (e.g., high-speed motion pushing the platform to its physical limits) and strategy (e.g., overtaking or blocking) emerge from agents trained with reinforcement learning.
We provide evidence in both simulation and the real world that this approach outperforms the common paradigm of training agents in isolation with rewards that prescribe behavior, e.g., progress on the raceline, in particular when the complexity of the environment increases, e.g., in the presence of obstacles.
Moreover, we find that multi-agent competition yields policies that transfer more reliably to the real world than policies trained with a single-agent progress-based reward, despite the two methods using the same simulation environment, randomization strategy, and hardware. 
In addition to improved sim-to-real transfer, the multi-agent policies also exhibit some degree of generalization to opponents unseen at training time.
Overall, our work, following in the tradition of multi-agent competitive game-play in digital domains, shows that sparse task-level rewards are sufficient for training agents capable of advanced low-level control in the physical world. \href{https://github.com/Jirl-upenn/AgileFlight_MultiAgent}
     {\textcolor{blue!60!black}{\faGithub\enspace{Code}}} ~\href{https://youtu.be/AIUfCbEJX6E}
     {\textcolor{blue!60!black}{\faYoutube\enspace{Video}}}

\end{abstract}

\section{INTRODUCTION}

Drone racing, a competitive sport where pilots fly quadcopters through tortuous circuits at high speed, has become a widely adopted benchmark for autonomous control~\cite{hanover2024autonomous}. Indeed, its requirement for decision-making under tight time constraints with little tolerance for errors makes it an ideal setting for testing advanced control strategies. In recent years, reinforcement learning (RL) has shown remarkable success on this task, consistently outperforming classical optimal-control techniques and even rivaling champion-level human pilots~\cite{Song23Reaching,kaufmann23champion,ferede2025one}. RL’s strength lies in its ability to optimize over long horizons and high-dimensional inputs. Such capabilities are difficult to achieve with model-based controllers, given their heavy reliance on online optimization.

\begin{figure}[t]
    \centering
    \includegraphics[width=0.9\linewidth]{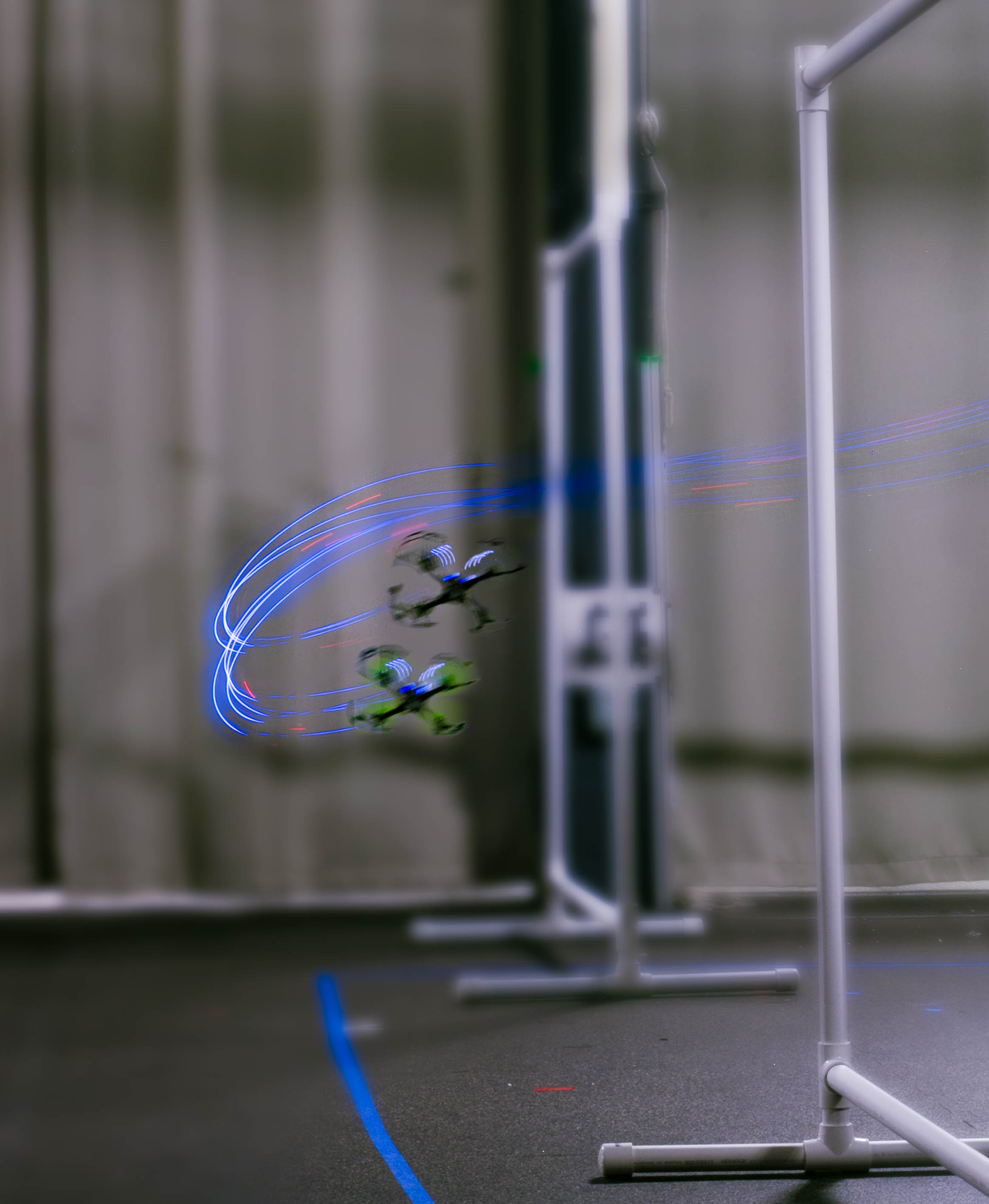}
    \caption{Two opponent-aware quadrotors performing head-to-head autonomous racing. The multi-agent policies are trained in simulation with a competitive, sparse, task-level reward (i.e., winning the race), without any specific behavior-based reward (e.g., flying fast), and are transferred zero-shot on the real-world-drones.}
    \label{fig:fig1_photo}
\end{figure}

Despite this promise, most RL approaches for drone racing remain more closely tied to optimal control than is often acknowledged. In particular, they typically require dense shaping rewards such as progress on the segment connecting two consecutive gates~\cite{song2021autonomous,Song23Reaching,kaufmann23champion,ferede2025one,kang2024autonomous}, which closely resembles trajectory-tracking costs used in optimal controllers~\cite{romero2021model,romero2022replanningRAL,foehn2021time}. As a result, most prior work trains RL agents to effectively follow a race line as fast as possible. This strategy prioritizes speed but prescribes behavior in a way that is not fundamentally different from trajectory optimization. Yet, as insights from game theory suggest~\cite{spica2018game,pustilnik2025non,zhu2024sequential}, maximizing speed alone is not always the optimal strategy for winning a race. Successful racing also requires tactical behaviors such as overtaking, blocking, or avoiding collisions, which are not easily captured by progress-based costs.

This observation raises a natural question: \emph{can RL agents learn racing strategies directly from outcome-based objectives, such as winning, without relying on dense behavioral shaping?} Our key insight is that this is possible when drone racing is framed as a multi-agent problem. By rewarding agents only for finishing a lap before their opponent, we show that fast and \emph{agile flight emerges naturally from sparse, competition-based rewards}. Moreover, this formulation gives rise to sophisticated behaviors, \emph{e.g.}, overtaking, blocking, and collision avoidance, despite the absence of any explicit reward terms for these skills. In contrast, we find that dense progress-based rewards, even when supplemented with overtaking terms~\cite{lee2025champion,kang2024autonomous}, are limiting as the complexity of the track increases, \emph{e.g.}, in the presence of obstacles, because their prescriptive structure constrains exploration during training.

Our work builds on the tradition of learning from multi-agent interaction~\cite{sims94creature,jaderberg2019human,liu2019emergent,cusumanorobust,vinitsky2023learning}, where rich behaviors emerge from simple task-level competitive rewards. However, in contrast to prior studies that primarily examined abstract or simplified environments, we observe a similar effect, albeit at a smaller scale, in a physically realistic, embodied setting, where learned policies can be directly deployed to real-world drones. Interestingly, we find that policies trained with sparse multi-agent rewards transfer more reliably to the real world than the ones trained with dense progress rewards, despite the two methods sharing the same simulation environment, randomization strategy, policy structure, and physical hardware.
With our work, we aim to inspire a shift in perspective within the real-world control community: from designing controllers that prescribe specific behaviors to designing controllers that optimize task-level objectives, letting desired behaviors emerge naturally.

\noindent \textbf{Contributions}. (1) We show that formulating autonomous drone racing as a competitive multi-agent problem naturally induces agile flight and tactical behaviors without explicit behavioral shaping. (2) We demonstrate that this approach outperforms dense progress-based rewards, in particular as the track complexity increases, while transferring better to the real world, and (3) We show generalization of our policies to agents it was not trained on.
    
\section{RELATED WORK}

\subsection{Single-Agent Drone Racing}
A wide range of controllers has been explored to tackle drone racing, ranging from classical methods to approaches based on deep learning~\cite{hanover2024autonomous}.

Among classical approaches, Model Predictive Control (MPC) and its variants are by far the most widely adopted. In~\cite{foehn2021time}, an MPC controller tracks a time-optimal trajectory computed offline. 
However, disturbances may lead the controller to suboptimal performance. To reduce this sensitivity, Model Predictive Contouring Control (MPCC) methods~\cite{romero2021model, romero2022replanningRAL} perform online adaptation of the path, velocities, and accelerations based on a track reference computed offline, leading to greater robustness.

While effective, the previous approaches face two main challenges: they require an accurate drone model, which is difficult to obtain, especially when considering aerodynamic effects during complex maneuvers; also, they often require significant computation power at inference time, which requires them to optimize relatively short-horizon objectives.
A natural solution to these challenges is provided by RL-based controllers~\cite{ferede2023end,ferede2024end, ferede2025one,song2021autonomous,Song23Reaching}. This approach waives the requirement of a very accurate model~\cite{zhang2025learning} and removes strict separation between long-term planning and control.

However, these methods focus on the performance of a single drone and do not study the dynamics of interaction between multiple agents. In addition, they require carefully designed reward functions to obtain good performance. In this work, we show that these two issues are related: considering adversarial interaction directly leads to simplified reward shaping.

\subsection{Multi-Agent Drone Racing}

The literature on multi-drone autonomous racing is much sparser than that on single-agent racing. Existing approaches rely on different strategies. To model the interaction and compute the best response, some methods rely on MPC ~\cite{zhao2024gate} or game-theoretic formulations~\cite{spica2018game, spica2020real, di2023cooperative}. However, these methods face severe scalability issues, as online computational cost increases with the complexity of the environment, which limits them to relatively low-speed racing.

Reinforcement learning offers an alternative way to address these limitations. Prior works trained multi-agent policies by augmenting single-agent rewards with an additional `overtaking' term~\cite{kang2024autonomous,wang2024dashing}, following approaches used in multi-agent car racing for Gran Turismo~\cite{lee2025champion}. In contrast, we show that dense single-agent rewards are not only unnecessary—since agile flight naturally emerges from the competitive dynamics of racing—but can even degrade performance as track complexity increases (e.g., with obstacles).

Conceptually, our work is closely related to research on multi-agent games, such as Capture the Flag~\cite{jaderberg2019human}, StarCraft~\cite{vinyals2019grandmaster}, hide-and-seek~\cite{baker2019emergent}, and autonomous driving~\cite{vinitsky2023learning}. Similar to our approach, these studies demonstrate that training reinforcement learning agents with simple adversarial rewards at scale can lead to complex emergent behaviors, such as the use of tools~\cite{baker2019emergent} or the development of social norms~\cite{vinitsky2023learning}. However, these works primarily focus on high-level strategy and abstract away low-level dynamics, which confines them to simulation environments. In contrast, our work shows that interesting behaviors also emerge when training policies on realistic dynamical systems, enabling zero-shot transfer of the policies to real hardware.

\section{METHODOLOGY}
In the following, we provide details of the policy optimization procedure and describe the dynamics model used for training.

\begin{figure*}[t]
    \centering
    \includegraphics[width=\textwidth]{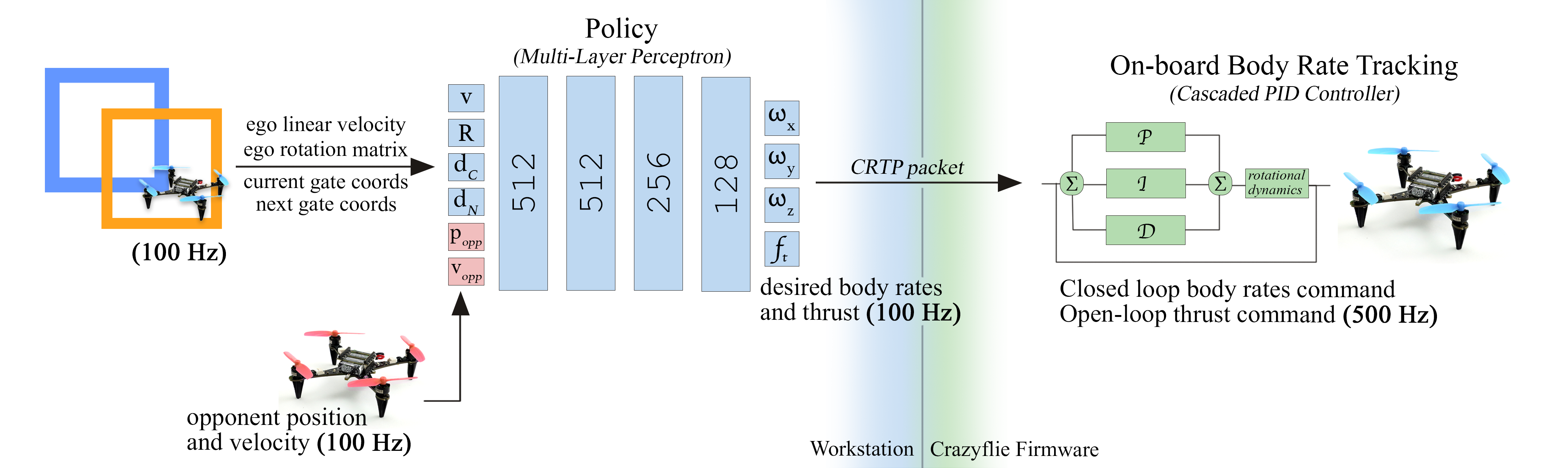}
    \caption{Observation to motor-command pipeline. Each drone's actor network receives ego-centric and opponent state estimates at 100 Hz from the Vicon motion capture system. A multi-layer perceptron processes the observations and outputs desired body rates and thrust, which is then sent to the drone via Crazy Real-Time Protocol (CRTP) at 100 Hz. Body rates are tracked via on-board rate PID, whereas thrust is sent as open-loop motor commands.}
    \label{fig:ActorMLPPID}
\end{figure*}

\subsection{Multi-Agent Policy Optimization}

We define drone racing as a multi-agent general-sum game between two agents $\pi^\E$ and $\pi^\A$, referred in the following as the \emph{\ego} and \emph{\adversary}. We jointly optimize the policies to maximize the following discrete-time finite-horizon objective:
\begin{equation}
\label{eq:objective}
    J(\pi^{e}, \pi^{a}) = \mathbb{E}_{\tau \sim p(\tau \mid \pi^\E, \pi^\A)} \Big[ \sum^{\thor}_{t=0} \gamma^t r_t(\xe_t, \xa_t) \Big],
\end{equation}
where $t$ denotes the discrete time step, $T$ represents the finite time horizon, and $r_t$ is a competitive sparse task reward. Here, $\tau = \{(\xe_t, \xa_t, \ue_t, \ua_t, r_t)\}^T_{t=0}$ is the joint trajectory of states, actions, and rewards induced by the \ego and \adversary policies, drawn from the distribution $p(\tau \mid \pi^\E, \pi^\A)$. The discount factor is denoted by $\gamma$. More formally, this optimization defines the solution to a two-agent, finite-horizon decentralized Markov decision process (Dec-MDP). 

We denote the agents global physical state as $\xe,\xa \in \mathbb{R}^{n}$ (with $n$ varying according to the environment). Similarly to prior work~\cite{kaufmann23champion, Song23Reaching}, the policies are reactive and predict the drones' vertical thrust and body rates, i.e., $\ua,\ue \in \mathcal{U} = [f_{z}^{des},\boldsymbol{\omega}_{des}]$ (see Sec.~\ref{sec:parametrization}). Such actions are then transformed to rotor forces with a fixed low-level PID controller (see Sec.~\ref{sec:simulation}).

\noindent \textbf{Rewards.} We use a sparse reward structure that encourages emergent racing behaviors without explicitly defining strategies: the agents are rewarded for passing a gate before their opponent ($r_t^{\text{pass}}$) and receive a bonus if completing a full lap first ($r_t^{\text{lap}}$). In addition to this sparse reward, we add an energy minimization term ($r_t^{\text{cmd}}$) as regularization, as well as a penalty for crashing into the ground or out of bounds ($r_t^{\text{crash}}$). Therefore, \begin{equation}
r_t = r_t^{\text{pass}} + r_t^{\text{lap}} - r_t^{\text{cmd}} - r_t^{\text{crash}}.
\end{equation}
Specifically, the terms are defined as follows:
\begin{align}
r_t^{\text{pass}} &= \begin{cases}
10.0, & \text{if gate passed at $t$ and leading} \\
5.0, & \text{elif passed gate at timestep} \\
0, & \text{otherwise}
\end{cases} \\
r_t^{\text{lap}} &= \begin{cases}
50.0, & \text{if lap completed at $t$ and leading} \\
0, & \text{otherwise}
\end{cases} \\
r_t^{\text{cmd}} &= -0.15 \, (\omega_{roll}^2 + \omega_{pitch}^2) -0.05 \, \omega_{yaw}^2 \\
r_t^{\text{crash}} &= \begin{cases}
2.0, & \text{if terminally crashed} \\
0.1, & \text{elif in contact} \\
0, & \text{otherwise}
\end{cases}
\end{align}
where $\omega_{\text{roll}}$, $\omega_{\text{pitch}}$, and $\omega_{\text{yaw}}$ are the commanded body rates. The reward scales are chosen empirically and we do not modify the environment in any way by adding extra ``phantom" gates to bias the drone trajectory.

\noindent \textbf{Comparison to Existing Rewards.} Our reward design removes a key component used in existing approaches: \emph{race-line progress}. This is typically defined as
\begin{equation}
\label{eq:progress}
    r_t^{\text{Prog}} = d^{\text{Gate}}_{t-1} - d_t^{\text{Gate}},
\end{equation}
where $d^{\text{Gate}}_t$ denotes the distance to the center of the next gate at time $t$. Intuitively, this dense reward encourages the agent to reach the next gate as quickly as possible while remaining close to the straight line connecting consecutive gates. However, we argue that such a formulation, reminiscent of reference trajectories in model-based controllers~\cite{romero2021model}, is overly restrictive. For example, it discourages deviations needed for obstacle avoidance and penalizes higher-level strategies such as blocking, overtaking, or adapting to an opponent’s crash. While existing methods have tried to overcome these issues in multi-agent settings by adding over\st{-}taking rewards~\cite{kang2024autonomous,wang2024dashing,lee2025champion}, we found this to be inferior to our approach, especially in the presence of obstacles.

\noindent \textbf{Optimization} We jointly train $\pi^\E$ and $\pi^\A$ to optimize Eq.~\ref{eq:objective} using IPPO~\cite{yu2022surprising}, a multi-agent variant of PPO~\cite{schulman2017proximal}. IPPO differs from standard PPO in that the surrogate objective is computed as the average of the individual agents' surrogate losses. Unlike MAPPO, IPPO does not employ a shared critic, which would require larger networks and potentially longer trainings. Instead, each agent maintains its own policy and critic, a design that we found to yield strong empirical performance. For this reason, and to keep the method simple, we did not explore alternative multi-agent RL formulations.

\subsection{Actor and Critic Parametrization}
\label{sec:parametrization}

\textbf{Actor Network}. Both agents are parametrized as MLPs with the same architecture and observation format. We use a 42-dimensional state vector defined as
\(\mathbf{x}^a_t, \mathbf{x}^e_t = [\mathbf{v}, \mathbf{R}, \mathbf{d}_C, \mathbf{d}_N, \mathbf{p}_{\text{opp}}, \mathbf{v}_{\text{opp}} \,]\),
where \(\mathbf{v} \in \mathbb{R}^3\) is the linear velocity of the drone expressed in body frame, 
\(\mathbf{R} \in \mathbb{R}^{3 \times 3}\) is the attitude rotation matrix, 
\(\mathbf{d}_C, \, \mathbf{d}_N \in \mathbb{R}^{12}\) are the position of the current and next gates' corners in body frame, respectively, and \(\mathbf{p}_{\text{opp}}, \mathbf{v}_{\text{opp}} \in \mathbb{R}^3\) are the opponent position and linear velocity in body frame, which are provided by a Vicon motion capture system.
In the tracks with obstacles, we extend the state space to $\mathbf{x}^a_t, \mathbf{x}^e_t,\in \mathbb{R}^{45}$ by including the global position \(\mathbf{p} \in \mathbb{R}^3\) of the drone.
Both policy networks have dimensions [512, 512, 256, 128] with ELU activation. 

\textbf{Critic Network}
For simplicity, we use separate critic networks for the two agents. Each critic receives privileged input in the form of the concatenated joint state, $x_{\text{critic}} = [\xe, \xa]$, and is implemented as a fully connected MLP. The hidden layers have dimensions [512, 512, 256, 256, 128, 128] with ELU activations. We empirically find that making the critic deeper than the actor improves performance. 

\subsection{Quadrotor Simulation}
\label{sec:simulation}

We train $\pi^\E$ and $\pi^\A$ exclusively in simulation and deploy them zero-shot to the real world. To keep the approach simple, we avoid extensive system identification and instead follow prior work~\cite{zhang2025learning,ferede2025one} by relying on domain randomization during training and rapid adaptation at test time. Our quadrotor dynamics model is based on the one in~\cite{kunapuli2025leveling}, with an additional aerodynamic component to capture drag, implemented following the model proposed in~\cite{forster2015system}.

Our quadrotor simulation, implemented in Isaac Sim~\cite{mittal2023orbit}, models a cascaded control architecture. A high-level controller, running at $50$ Hz, generates thrust and angular rate commands, while a low-level controller, running at $500$ Hz, tracks the desired body rates using a PID. These commands are converted into motor forces and torques and applied, together with aerodynamic effects, to the quadcopter, which is modeled as a rigid body.

Denote $[a_0, a_1, a_2, a_3]$ a policy output. The first element $a_0 \in [-1, 1]$ is mapped to the thrust along the body $z$-axis:
\begin{equation}
    f_z^{\text{des}} = \left( a_0 + 1 \right) / \ 2 \cdot T,
\end{equation}
where $T$ is the maximum thrust expressed in units of drone weight.
The remaining action components $a_1,a_2,a_3 \in [-1, 1]$ determine the desired angular velocities:
\begin{equation}
    \boldsymbol{\omega}^{\text{des}} =
    \begin{bmatrix}
        k_{x} a_1, \, k_{y} a_2, \, k_{z} a_3
    \end{bmatrix},
\end{equation}
where $k_{x}$, $k_{y}$, and $k_{z}$ are the maximum roll, pitch, and yaw rate, respectively. 
The angular velocity error $\boldsymbol{e}_\omega = \boldsymbol{\omega}^{\text{des}} - \boldsymbol{\omega}$ is then converted to desired torques via a PID control law, after scaling by the inertia matrix $\mathcal{I}$: 
\begin{equation}
    \boldsymbol{\tau}^{\text{des}} = \mathcal{I} \left( K_p \boldsymbol{e}_\omega - K_d \boldsymbol{\dot{e}}_\omega + K_I {\textstyle \int} \boldsymbol{e}_\omega \right).
\end{equation}

The desired wrench $\boldsymbol{w}^{\text{des}} = [f_z^{\text{des}}, \boldsymbol{\tau}^{\text{des}}]$ applied to the rigid body is then converted to motor forces by preconditioning with the inverse of the thrust-to-wrench static mapping, i.e., $\boldsymbol{f}_{\text{mot}}^{\text{des}} = T_M^{-T} \cdot \boldsymbol{w}^{\text{des}}$.
From these forces, the desired motor speeds for each motor $i = 1, \dots, 4$ are computed using the thrust coefficient $k_\eta$, i.e., $\boldsymbol{\omega}_{\text{mot}}^2 = \boldsymbol{f}_{\text{mot}}^{\text{des}} \, / \, k_\eta$. These are then converted to actual motor speeds using the minimum and maximum allowable motor speeds, $\underline{\omega}$ and $\bar{\omega}$:
\begin{equation}
    \omega_{\text{mot, i}}^{\text{des}} = \omega_{\text{i}}^{\text{des}} = \text{clamp}\left( \sqrt{|\omega_i^2|} \cdot \text{sign}(\omega_i), \underline{\omega}, \bar{\omega} \right).
\end{equation}
We model the motor dynamics with a first-order model governed by the motor constant $\tau_m$:
\begin{equation}
    \dot{\omega}_i = \frac{\omega_i^{\text{des}} - \omega_i}{\tau_m}, \quad
    \omega_i \leftarrow \omega_i + \dot{\omega}_i \cdot \Delta t.
\end{equation}
Once we compute the actual motor speed, we compute the actual forces and wrench applied to the body as
\begin{equation}
    f_i = k_\eta \cdot \omega_i^2, \quad \boldsymbol{w} = T_M \cdot \boldsymbol{f} =
    \begin{bmatrix}
        f_z, \boldsymbol{\tau}^{T}
    \end{bmatrix}^{T}.
\end{equation}

In addition, similarly to~\cite{forster2015system}, we model aerodynamic effects as forces and torques proportional to the translational and angular velocities:
\begin{equation}
    \boldsymbol{f}^{\text{drag}} = - \sum \omega_i \, \boldsymbol{K}_{\text{aero}} \, \boldsymbol{v}_b \,
\end{equation}
where $\boldsymbol{K}_{\text{aero}}$ is a diagonal matrix, $\boldsymbol{v}_b$ is the linear velocity of the drone expressed in the body frame.
The final applied force and torque acting on the rigid body combine the motor forces and aerodynamics, and are:
\begin{equation}
    \boldsymbol{F}_{\text{app}} =
    \begin{bmatrix}
        f_x^\text{drag}, f_y^\text{drag}, f_z^\text{drag} + f_z
    \end{bmatrix}^T, \quad
    \boldsymbol{\tau}_{\text{app}} = \boldsymbol{\tau}.
\end{equation}

\section{Experiments}
Our evaluation is designed to find answers to the following questions. What are the limitations of single-agent racing rewards? How do agents perform in head-to-head races? Can multi-agent policies bridge the gap between controlled experiments and deployment on physical platforms? Does a qualitative analysis reveal the emergence of strategic behaviors? Finally, we study the stability of training multi-agent policies with sparse rewards.

\subsection{Experimental Setup}

\begin{figure*}[t]
    \centering
    \begin{subfigure}[b]{0.48\textwidth}
        \centering
        \includegraphics[width=\linewidth]{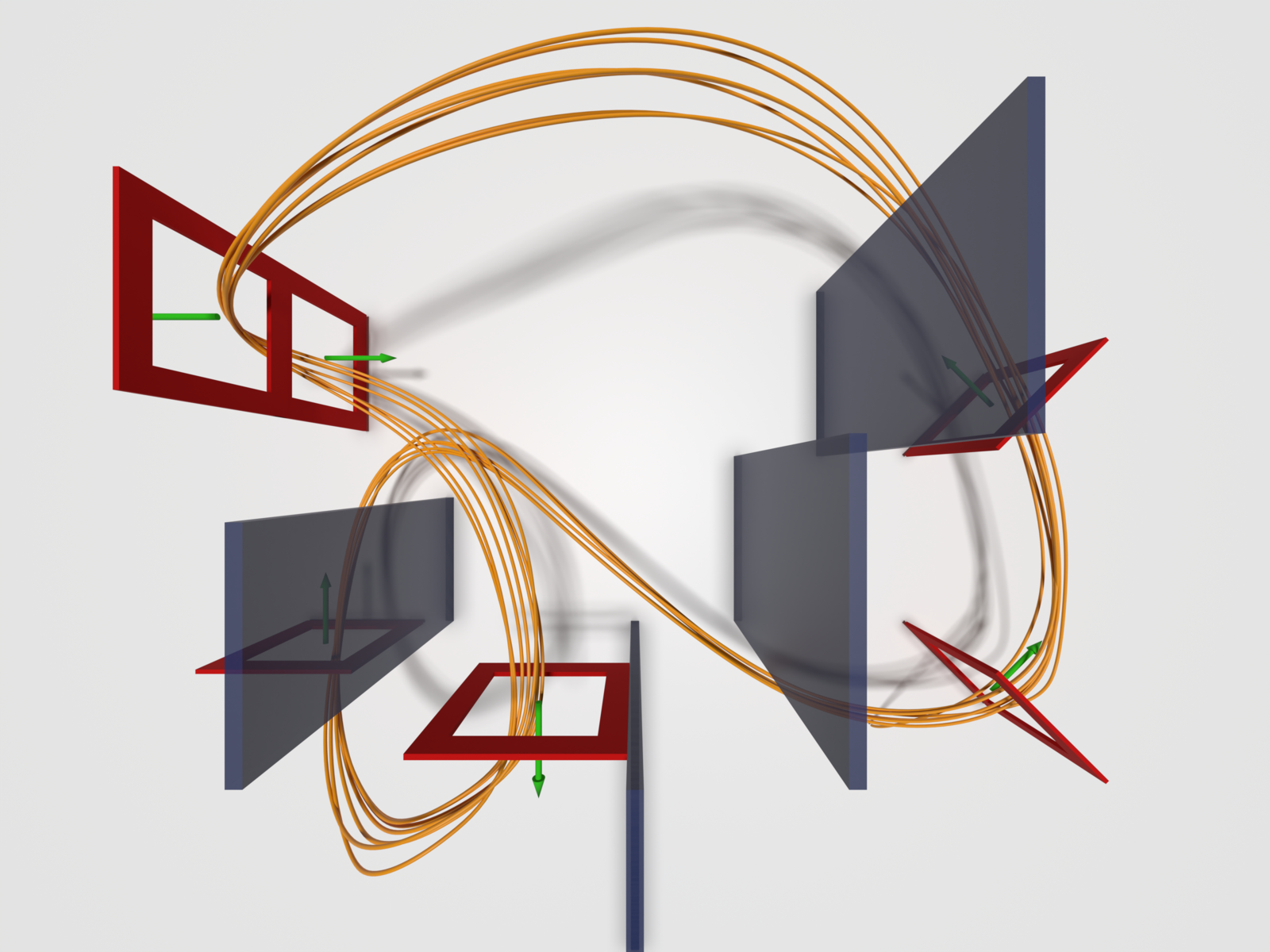}
    \end{subfigure}
    \hfill
    \begin{subfigure}[b]{0.48\textwidth}
        \centering
        \includegraphics[width=\linewidth]{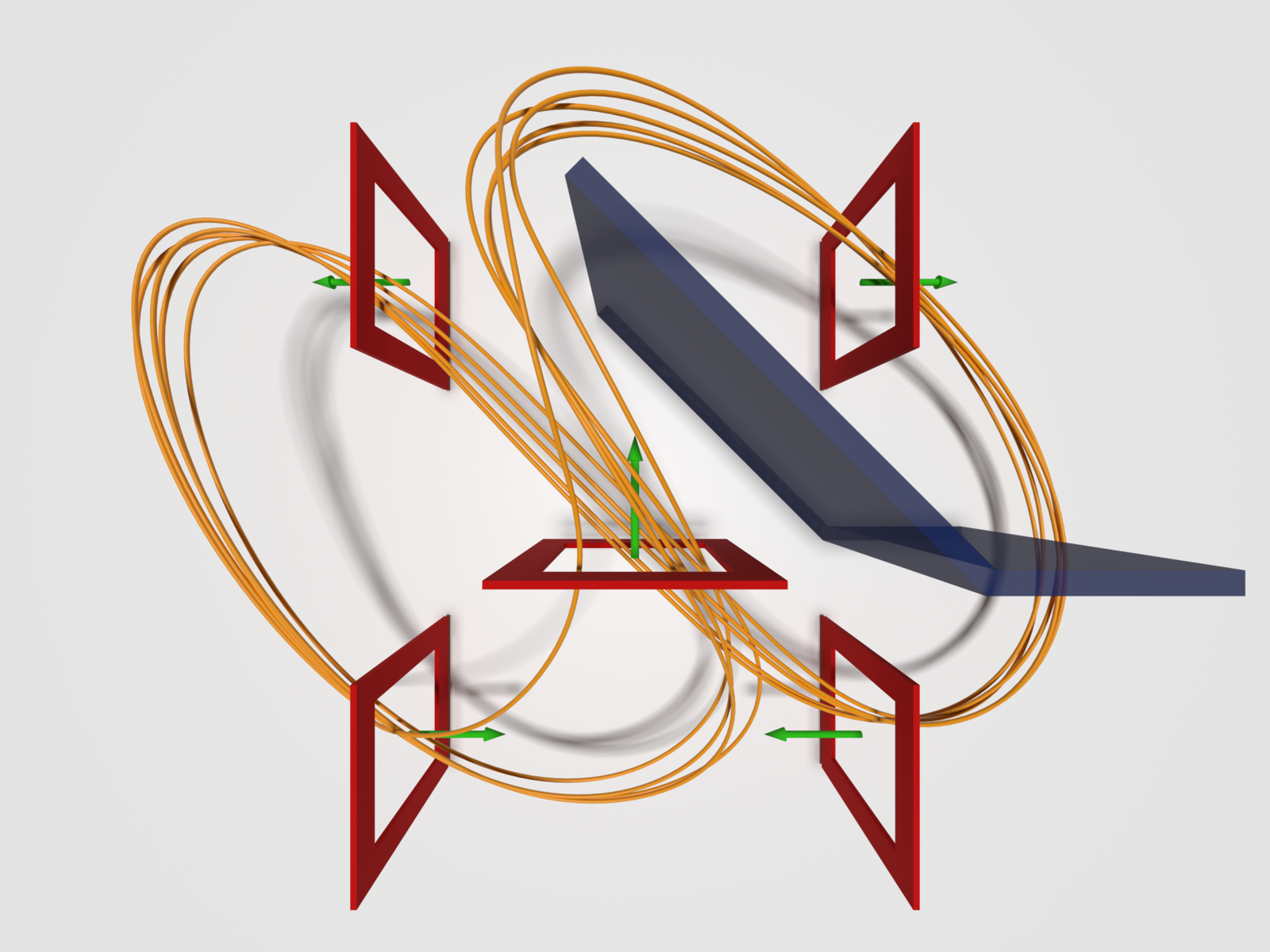}
    \end{subfigure}
    \caption{Tracks used for both training and evaluation, with $1~\mathrm{m} \times 1~\mathrm{m}$ gates shown in red and obstacles in transparent blue. Green arrows indicate the gate-passing directions, and the orange curves show the trajectories followed by our drone over multiple laps in simulation. On the left, the Complex Track (CT) spans $8~\mathrm{m} \times 7~\mathrm{m}$ and features six gates, including a slit-S, and optionally four obstacles. On the right, the Lemniscate Track (LT) measures $5~\mathrm{m} \times 5~\mathrm{m}$, with five gates—one of which is passed twice in a single lap—and optionally two overlapping obstacles}
    \label{fig:complex_lemniscate}
\end{figure*}

For the simulation, we utilize Isaac Lab v2.2.0 and Isaac Sim v4.5.0 \cite{mittal2023orbit}, where we developed a custom environment with a Crazyflie 2.1 Brushless. This quadrotor was chosen due to its compact form factor ($3''$), lightweight design (32~g), and high thrust-to-weight ratio (slightly greater than~3). Two tracks were designed for the experiments: the \textit{Complex Track} (CT) and the \textit{Lemniscate Track} (LT), shown in Fig.~\ref{fig:complex_lemniscate}, along with the gate-passing directions, obstacles, and a drone trajectory. The CT consists of six gates, including a split-S maneuver, and four obstacles, while the LT features five gates, one of which is crossed twice per lap, and two obstacles. Both track layouts were designed to be challenging enough to require agile motions, even in the absence of obstacles. Additionally, when obstacles are present, their placement forces the drones to first move away from the next gate before reaching it.
In the real-world setup, flights were carried out within a controlled indoor area, measuring $22.0~\text{m} \times 5.5~\text{m} \times 3.8~\text{m}$. A motion capture system with 25 Vicon cameras provided ground-truth poses at a frequency above 100~Hz. 

We benchmark against four policies that differ in the training setup: (Single- or Multi-Agent) and in the reward type (dense or sparse). These policies are denoted as \textit{Dense Single-Agent} (DS), similar to~\cite{song2021autonomous, Song23Reaching, kaufmann23champion, ferede2025one, kang2024autonomous}; \textit{Sparse Single-Agent} (SS), which uses the same sparse reward as ours but is not trained with a competitor; \textit{Dense Multi-Agent} (DM), which, similarly to~\cite{lee2025champion,kang2024autonomous} complements the progress reward with the dense overtaking reward
\begin{equation*}
    r_t^{\text{ot}} = \mathbb{I}_{d_{min} < ({p}^{i}_{t} - p_{t}) < d_{max}} \big( (p_{t} - {p}^{i}_{t}) - (p_{t-1} - {p}^{i}_{t-1}) \big), 
\end{equation*}
where $d_{min} = -5$ m and $d_{max} = 10$ m; 
and \textit{Sparse Multi-Agent} (Ours).
In the multi-agent cases, we evaluate the actor that received the greatest total reward during training and discard the other.
All baselines are carefully tuned to obtain the best performance.

To assess the performance of each policy, we performed two different types of tests. The first test evaluated the single-agent policies in simulation by varying the track layout and the presence of obstacles. The agents were not directly provided with the obstacle positions; instead, the algorithm inferred them by exploiting the global drone positions. As a result, eight independent experiments were conducted, each consisting of fifty separate 3-lap runs with different initial positions beyond a selected gate. Four metrics were used for evaluation: the average Time To Complete (TTC) one lap in seconds, the average speed in meters per second (both computed over the fifty runs), the total number of collisions per experiment against gates and obstacles, and the success rate, measured as the proportion of 3-lap runs successfully completed out of fifty trials.

The second set of tests focused on head-to-head races between all pairs of trained policies across all the track layouts. Additionally, several multi-agent experiments were conducted in the real world using the lemniscate track. In both cases, the primary metric was the win rate of one policy against the other, expressed as a percentage of the total number of races, where each race consists of 3 laps. We do fifty races in simulation and three in the real world. The first agent to complete three laps is considered the winner, while a terminal crash by both agents is counted as a draw.

\subsection{What are the limits of single-agent racing?}

\begin{table}[b]
    \centering
    \resizebox{\columnwidth}{!}{%
    \begin{tabular}{lccccc}
        \toprule
              & Avg. TTC [\si{\second}] & Avg. Speed [\si{\meter\per\second}] & Collisions & Success [\%]& \\ 
        \midrule
        LT        & 4.76  & 4.07 & 0   & 100 & \multirow{4}{*}{\rotatebox{90}{\textbf{Dense}}} \\
        LT w/ obs & --    & --   & 175 & 0  &  \\
        CT        & 5.68  & 5.01 & 1   & 98 &  \\
        CT w/ obs & --    & --   & 0   & 0  &  \\
        \midrule
        LT        & 5.29  & 3.57 & 0   & 100 & \multirow{4}{*}{\rotatebox{90}{\textbf{Sparse}}} \\
        LT w/ obs & 5.72  & 3.95 & 12  & 98 &  \\
        CT        & 6.42  & 4.30 & 16  & 94 &  \\
        CT w/ obs & 10.08 & 4.58 & 69  & 0  &  \\
        \bottomrule
    \end{tabular}
    }
    \caption{Results of the single-agent simulated tests with different track configurations, reward functions, and obstacle presence indicate clear preference for dense rewards in obstacle-free environments.}
    \label{tab:dense_sparse}
    \vspace{-15pt}
\end{table}

The results of the policies trained with a single-agent formulation are presented in Table~\ref{tab:dense_sparse}. Notably, the policies trained with dense rewards achieved strong performance only in environments without obstacles, with a success rate of 100\% on the lemniscate track and 98\% on the complex track. In contrast, performance with obstacles is extremely poor, yielding a success rate of zero on both tracks, even failing to complete a single lap. This behavior is inherently due to Eq.~\ref{eq:progress}, the progress reward, which discourages the drone from moving away from the gate, preventing it from successfully overcoming obstacles.

\begin{figure*}[!t]
    \centering
    \includegraphics[width=\textwidth]{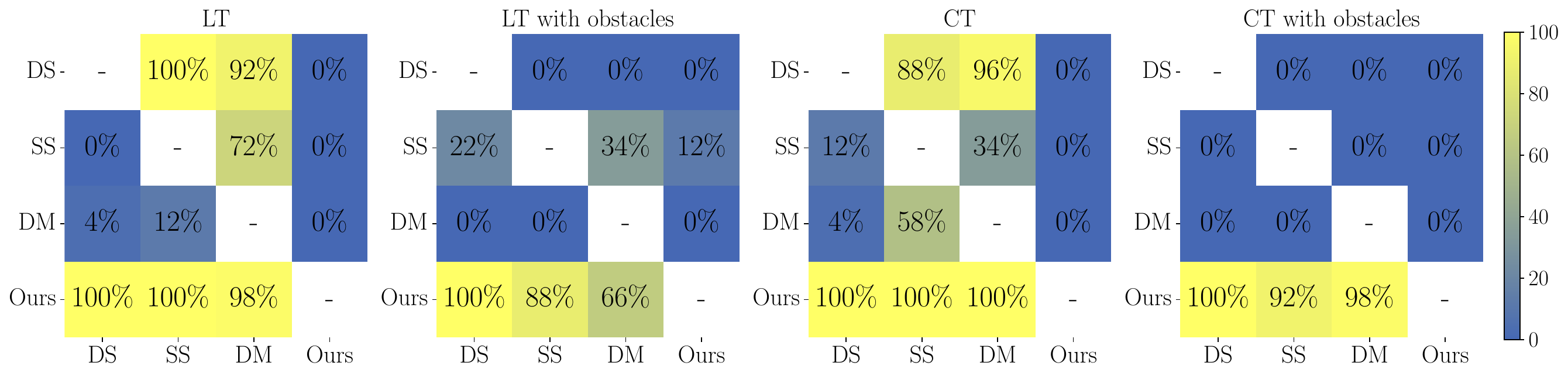}
    \caption{Head-to-head results in simulation by varying the track layouts and the racing policies. The latter are defined as Dense Single-agent (DS)~\cite{song2021autonomous, Song23Reaching, kaufmann23champion, ferede2025one, kang2024autonomous}, Sparse Single-Agent (SS), an ablation of our approach without competition, Dense Multi-Agent (DM), similar to DS with a competitive reward, and Sparse Multi-Agent (Ours). Each cell shows the win rate of the row policy against the column policy. The results of these experiments confirm that optimizing task-level objectives in a competitive manner not only waves the needs of dense behavioral rewards, but leads to overall better performance and flexibility.}
    \label{fig:h2h_result}
\end{figure*}

As for the sparse reward policies, their success rate on obstacle-free tracks is at times comparable to that of the dense policies, although with lower speed. In the lemniscate track, the average lap time of the SS policy is 5.29 s, compared to 4.76 s for DS, corresponding to an increase of 11.13\%, while the average speed drops from 4.07 \si{\meter\per\second} to 3.57 \si{\meter\per\second} (-12.3\%). In the complex track, the results are similar, with the lap time increasing from 5.68 s to 6.42 s (+13.0\%) and the speed decreasing from 5.01 to 4.30 \si{\meter\per\second} (-14.17\%). Additionally, applying the sparse reward on the lemniscate track with obstacles provides an almost perfect success rate of 98\%, whereas on the complex track the success rate drops to 0\%, despite the completion of a few individual laps.

Overall, these experiments show that in single-agent drone racing environments, dense rewards more reliably converge to competitive policies than sparse rewards, confirming the finding of prior work~\cite{Song23Reaching,kaufmann23champion}. Neither sparse nor dense rewards, however, consistently discover obstacle-free trajectories across tracks. Additionally, dense rewards are challenging to tune as the track complexity increases, e.g., in the presence of obstacles. While for static obstacles (as in our experiments) one could circumvent this problem by first doing coarse path planning and tracking the planned path, this solution is not general. For example, this heuristic would fail if the track requires avoidance of dynamic obstacles.

\subsection{Baseline comparison in simulated head-to-head races}

We evaluated our multi-agent policy's competitive performance through head-to-head races, as it provides a clearer indication of dominance than lap times alone. We pitted two drones against each other in every race, running one of the four methods. We use the same single-agent policies that were evaluated in the previous section.

The results of this evaluation are shown in Fig.~\ref{fig:h2h_result}, where each cell in the four matrices reports the win rate of the row policy against the column policy. The entries are symmetric with respect to the main diagonal, corresponding to the same policy pair with reversed roles. Note that these percentages do not sum to 100\%, as the remainder corresponds to the fraction of draws due to collision. We omit elements on the diagonal, as we don't evaluate a policy against itself.

The results of these experiments confirm that in a single-agent environment, a simple sparse reward does not frequently outperform a dense reward. 
The only scenario in which the SS policy shows a slight advantage is in the presence of obstacles, where it occasionally wins against DS. Extending the DS formulation to a multi-agent approach (DM) by adding a competition reward does not improve performance; in fact, it often results in more defeats compared to DS and still fails in the presence of obstacles. This outcome likely stems from the difficulty of balancing two dense reward terms, which could cause the policy to converge to a suboptimal local minimum. 

\begin{figure*}[t]
    \centering
    \includegraphics[width=\textwidth]{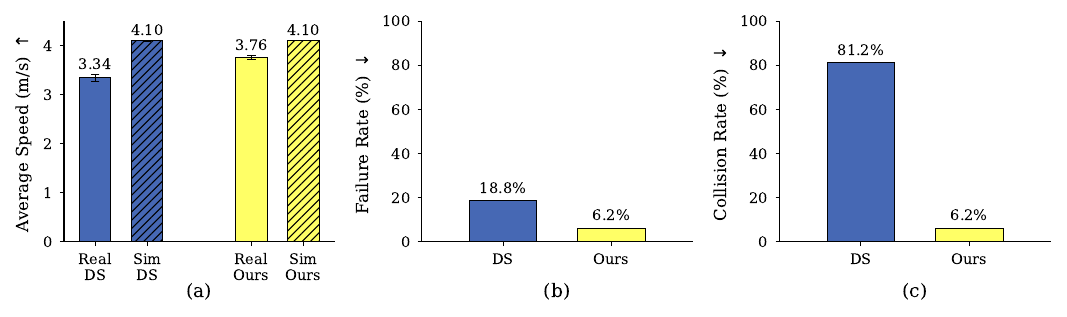}
    \caption{Evaluation of simulation-to-reality transfer for DS and Ours. Both methods use the same simulation environment and randomization strategy, and are deployed on identical hardware. Subplot (a) compares average in-flight speed between simulation and real-world deployment on the lemniscate track; subplot (b) computes failure rate: \textit{ terminal collisions / attempted laps} and (c) collision rate: \textit{ collisions / attempted laps}. Note that in simulation, both the failure and collision rates are zero for both methods across all tracks. However, our approach exhibits superior sim-to-real transfer, as evidenced by a smaller gap between the average speed in simulation and in the real world, along with lower real-world failure and collision rates.}
    \label{fig:sim2real_ds_ours}
\end{figure*}

Conversely, training with sparse competitive rewards yields the highest average win rates of 91.17\%, as shown by the performance of Ours. Interestingly, we outperform the dense baselines even when they are at their best in the tracks without obstacles. Specifically, we win 100\% (50/50) of the races on the lemniscate and 84\% (42/50) on the complex against DS. Occasionally, however, the win rate drops (62\% and 66\% against SS and DM, respectively, on the lemniscate track). This is often due to one of the agents colliding into our policy or behaving in an erratic way that our policy did not observe at training time. The success of Ours on obstacle-laden tracks indicates how competitive pressure among agents broadens the range of behaviors discovered during training, which is made especially clear when comparing against SS, which differs only in the number of agents.

Overall, the results of these experiments confirm the central hypothesis of this work: optimizing task-level objectives in a competitive manner not only waves the needs of dense behavioral rewards, but leads to overall better performance and flexibility.

\subsection{Zero-Shot Deployment to the Real World}

In this section, we evaluate how policies transfer to the real world. We test policies in a real-world replica of both the lemniscate and the complex track. We invite the reader to watch the supplementary video for more details.

We begin by comparing the simulation-to-real-world transfer performance of the DS policy and our method. For each track layout, we conduct three head-to-head races, each consisting of three laps. As shown in Fig.~\ref{fig:sim2real_ds_ours}, our approach achieves a 44.7\% smaller gap between the average flight speed in simulation and in the real world (0.76\,m/s for DS vs.\ 0.43\,m/s for ours), along with a substantially lower real-world failure and collision rate (in simulation, both rates are zero for both methods). These results indicate that competition with sparse rewards leads to improved sim-to-real transfer.

Qualitatively, we observe the same trend as the track complexity increases: our method is the only one that successfully completes the lemniscate track in the presence of obstacles. We believe this improved sim-to-real transfer to be related to adversarial domain randomization~\cite{khirodkar2018adversarial}, although we leave further analysis of this effect to future work.

\begin{figure}
    \centering
    \includegraphics[width=0.6\linewidth]{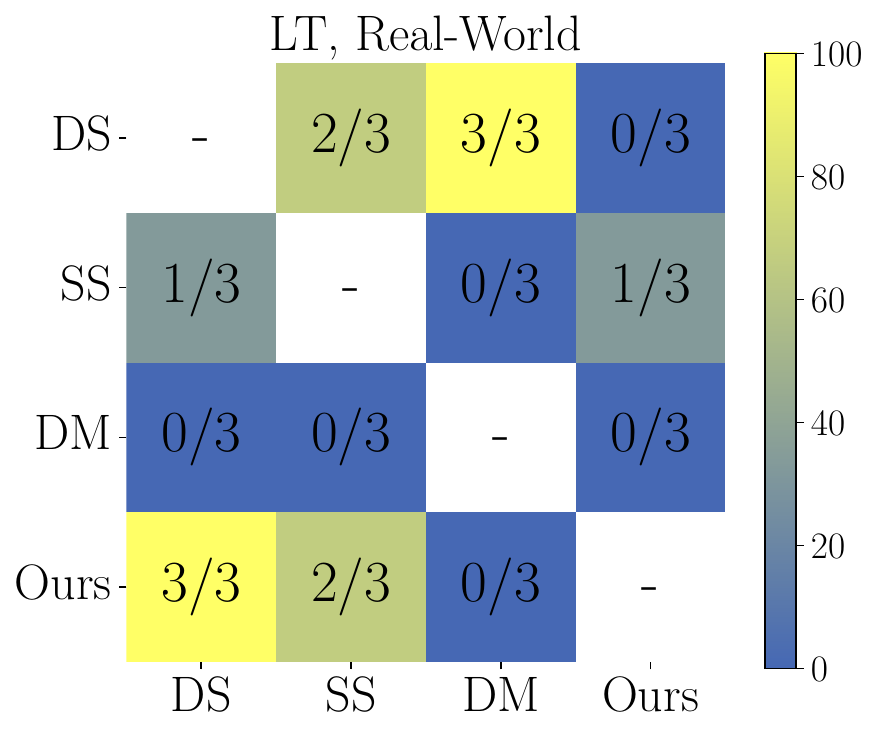}
    \caption{Real-world head-to-head results on the lemniscate track. DS and Ours achieve the same average win rate, but Ours outperforms DS in head-to-head races. This is because DM does not transfer well, and often exhibits behaviors that put our policy out of distribution.}
    \label{fig:h2h_results_real}
\end{figure}

We now evaluate the real-world robustness of our approach against opponents it was not trained on. To this end, we conduct head-to-head competitions among all methods on the lemniscate track. Fig.~\ref{fig:h2h_results_real} reports the results of these experiments, with each comparison conducted over three races. Our policy generalizes well across different opponents, achieving the highest win rate (on par with DS).
Additionally, we pit Ours in a real-world head-to-head race against DS on the complex track; Ours won 2 of 3 races and the third resulted in a draw due to collision.

\begin{figure*}[t]
    \centering
    \includegraphics[width=\textwidth]{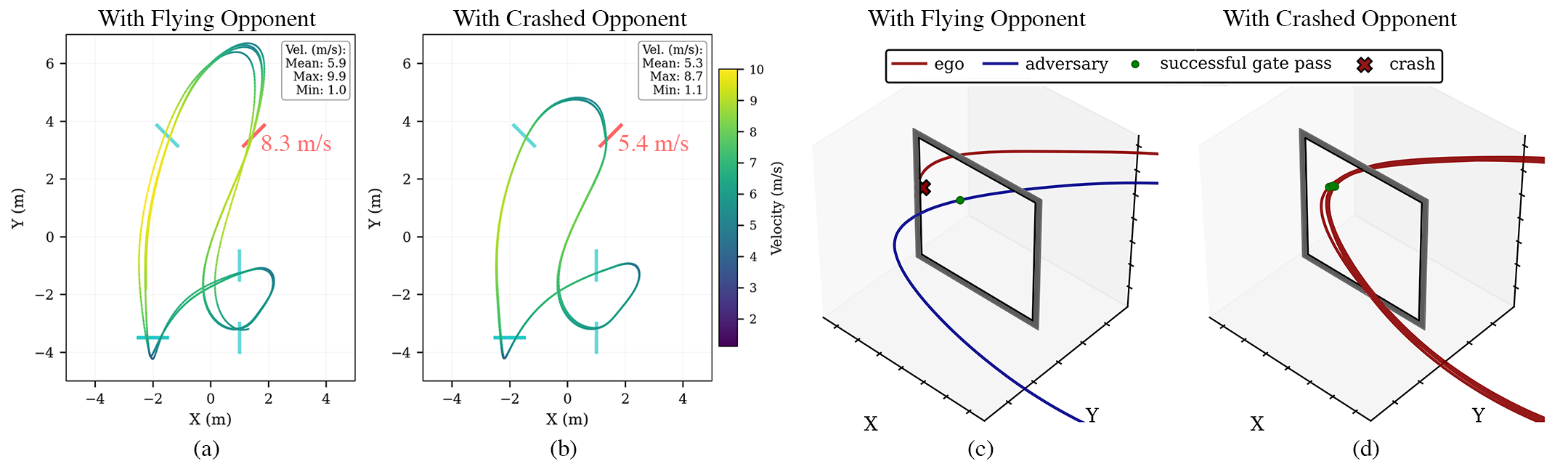}
    \caption{Subplots (a) and (b) respectively depict our agent's trajectory and velocity profile in head-to-head races against an adversary (not pictured) and a crashed drone. Our policy shows risk-averse behavior once the opponent is non-competitive. Subplot (c) depicts a learned blocking maneuver in a real-world race, and (d) illustrates a real-world trajectory flown against a crashed opponent as a comparison.}
    \label{fig:sm_v_sm_result}
\end{figure*}

With these experiments we make a subtle but interesting finding: Ours and DS have the same average win rates in the lemniscate track (Fig.~\ref{fig:h2h_results_real}), despite Ours outperforming DS in direct head-to-head races. This is due to the fact that our approach had 3 draws with the DM baseline. Interestingly, this is not because the DM policy is particularly strong. Conversely, it did not transfer very well, and ended up consistently losing against all opponents. However, its behavior had an unexpected influence on our policy, making it crash twice even if it had a clear lead and once at the beginning of the race, while both policies were rushing into the central gate. 
This highlights the challenges of simulation to reality transfer in multi-agent settings: the dependence on the behavior of another agent makes it challenging to predict the performance of a policy when deployed against new opponents. While a robust solution to this problem is outside of the scope of this work, we predict that large randomization at training time and adaptation at test time could alleviate these issues.

\subsection{Emergence of strategic behaviors}
We evaluated whether strategic behaviors emerge from sparse competitive rewards by analyzing trajectory and velocity profiles in head-to-head self-play. On the Complex Track, we find that our agent flies a significantly more aggressive trajectory against a competitive opponent than against a crashed opponent. (Fig.~\ref{fig:sm_v_sm_result} a-b). 
Specifically, our agent reaches a maximum velocity of 9.9 m/s and mean velocity of 5.9 m/s when flying against an opponent, but reduces to a maximum velocity of 8.7 m/s and mean velocity of 5.3 m/s once the opponent has crashed. 
We also observe that our agent maintains a substantially higher final-gate velocity of 8.3 m/s when the opponent is competitive, as opposed to 5.4 m/s when the opponent is crashed. This behavior is risk averse, as the gate-pass rewards and lap bonus rewards become guaranteed once the opponent forfeits the racing task.

Blocking maneuvers are another indicator of rich opponent-aware strategies emerging from sparse competitive multi-agent rewards. Fig.~\ref{fig:sm_v_sm_result}(c and d) depict one such maneuver in the real world. During competition, the adversary defends its marginal forward lead by flying a wide trajectory as it enters the gate (Fig.~\ref{fig:sm_v_sm_result}c), forcing the ego agent to the outside and causing a collision with the gate frame. In contrast, Fig.~\ref{fig:sm_v_sm_result}d illustrates the trajectory flown with a crashed opponent, which is clearly a safer trajectory.

In summary, the evidence from this section shows that our agent learns strategic behaviors beyond flying fast.

\subsection{Training stability of multi-agent policies}

\begin{figure}[t]
    \centering
    \includegraphics[width=0.95\linewidth]{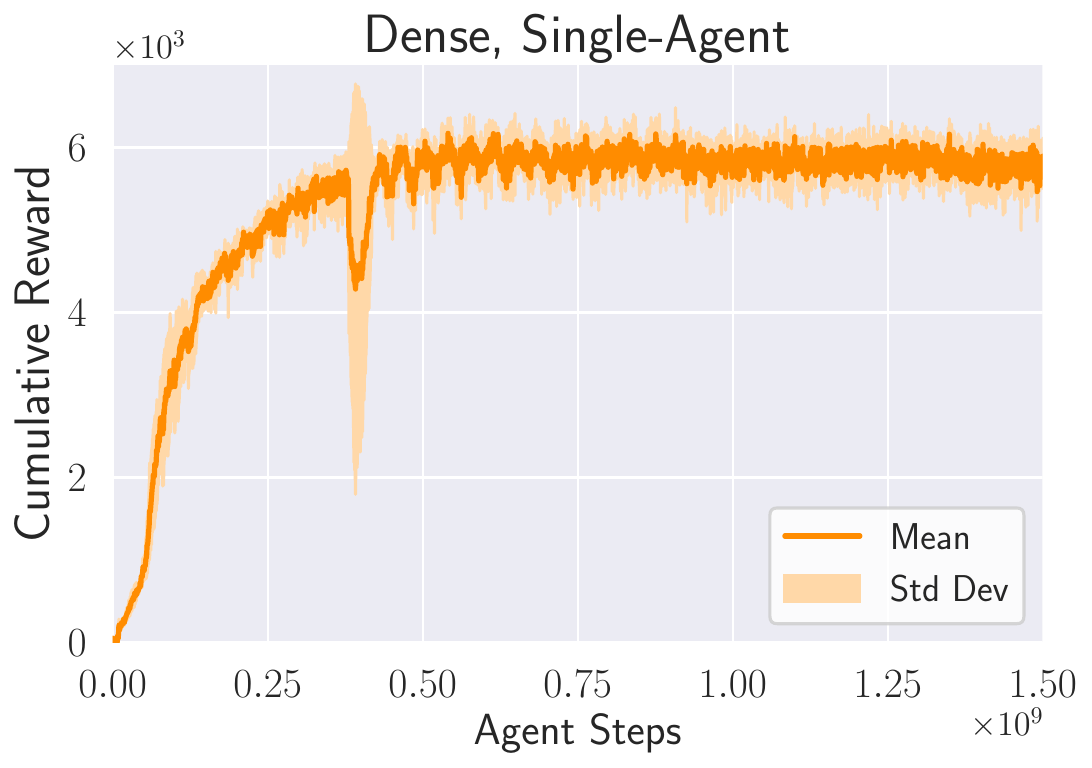}
    \caption{Cumulative reward for single-agent training. 
    The dark curves represent the mean rewards obtained across three different seeds. The training is more stable and smoother, with very low variance.}
    \label{fig:single_agent_cumulative_reward}
\end{figure}

\begin{figure}
    \centering
    \includegraphics[width=0.95\linewidth]{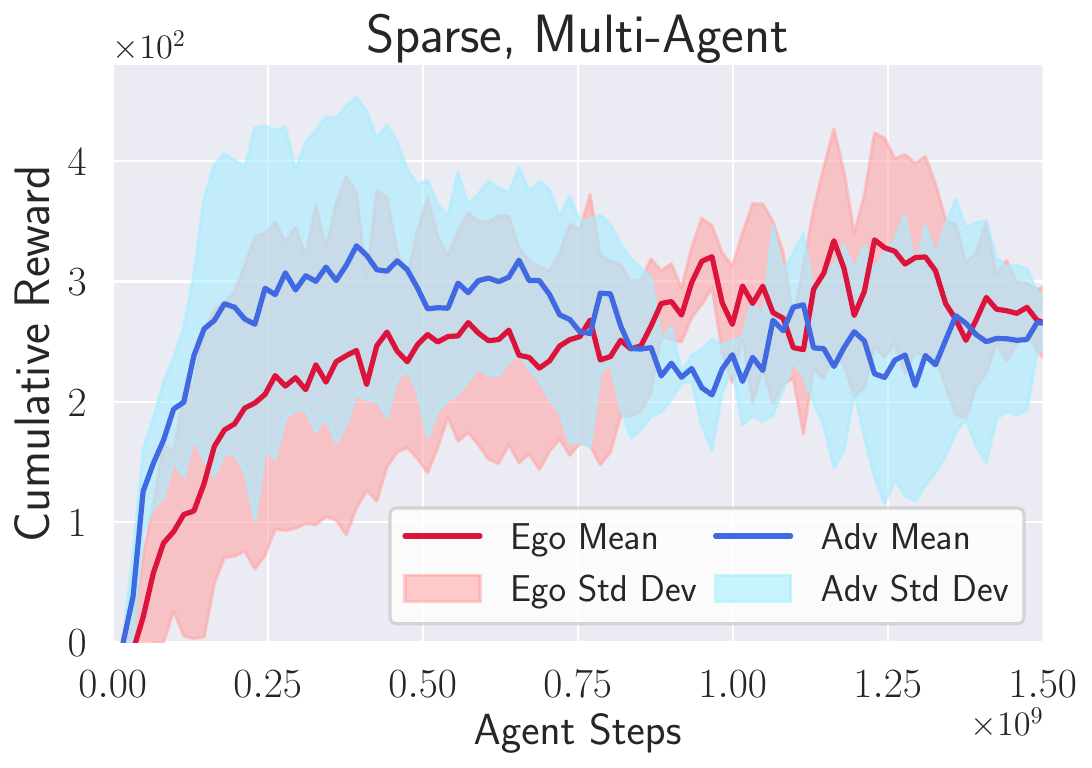}
    \caption{Cumulative reward for multi-agent training.  The training curves exhibit greater variability, yet remain consistent overall. The curves alternate in achieving higher cumulative rewards, reflecting the competitive nature of this setup.}
    \label{fig:multi_agent_cumulative_reward}
\end{figure}

The final analysis we conducted focused on the stability of training multi-agent policies, in comparison to the dense training of a single-agent policy. Fig.~\ref{fig:single_agent_cumulative_reward} and Fig.~\ref{fig:multi_agent_cumulative_reward} present the training curves for the DS policy and Ours. The plots report the mean and standard deviation across three independent runs, obtained with seeds 1, 2, and 3. As expected, the single-agent training is stable and smooth, with very low variance. In contrast, the multi-agent training curves exhibit greater variability, yet they remain consistent overall: all three runs train very competitive policies. Interestingly, the blue and red curves alternate in achieving higher cumulative rewards, reflecting the competitive nature of the multi-agent setup: as the two drones learn against each other, different phases emerge where one temporarily outperforms the other before the balance shifts again.

\section{CONCLUSIONS}

Our study provides evidence that agile and strategic low-level control in physical platforms can emerge from simple, sparse competition-based rewards.
Moreover, it highlights the advantages of such rewards over prescriptive terms that constrain agent behavior, both in terms of simulation to reality transfer and overall performance.
We believe, however, that this work only scratches the surface of what competitive rewards can achieve.
Future extensions could explore the scalability of this approach towards team-based competition, the emergence of active perception in vision-based agents, or the limits of these formulations against opponents that adapt rapidly to new strategies.

\section{ACKNOWLEDGMENTS}

This work was supported by the DARPA TIAMAT
program (HR0011-24-9-0430). We are grateful to the Google TPU Research Cloud Program, NSF ACCESS, and the National Center for Supercomputing Applications (NCSA) for providing computational resources.
We additionally would like to thank Himanshu Gaurav Singh, Pratik Kunapuli, and Chunwei Xing for helpful discussion and assistance throughout the project.

\addtolength{\textheight}{-12cm}  %

\def\UrlBreaks{\do\/\do-}
{\footnotesize
\bibliographystyle{IEEEtran}
\bibliography{references}

\begin{thebibliography}{10}
\providecommand{\url}[1]{#1}
\csname url@samestyle\endcsname
\providecommand{\newblock}{\relax}
\providecommand{\bibinfo}[2]{#2}
\providecommand{\BIBentrySTDinterwordspacing}{\spaceskip=0pt\relax}
\providecommand{\BIBentryALTinterwordstretchfactor}{4}
\providecommand{\BIBentryALTinterwordspacing}{\spaceskip=\fontdimen2\font plus
\BIBentryALTinterwordstretchfactor\fontdimen3\font minus \fontdimen4\font\relax}
\providecommand{\BIBforeignlanguage}[2]{{%
\expandafter\ifx\csname l@#1\endcsname\relax
\typeout{** WARNING: IEEEtran.bst: No hyphenation pattern has been}%
\typeout{** loaded for the language `#1'. Using the pattern for}%
\typeout{** the default language instead.}%
\else
\language=\csname l@#1\endcsname
\fi
#2}}
\providecommand{\BIBdecl}{\relax}
\BIBdecl

\bibitem{hanover2024autonomous}
D.~Hanover, A.~Loquercio, L.~Bauersfeld, A.~Romero, R.~Penicka, Y.~Song, G.~Cioffi, E.~Kaufmann, and D.~Scaramuzza, ``Autonomous drone racing: A survey,'' \emph{IEEE Transactions on Robotics}, vol.~40, pp. 3044--3067, 2024.

\bibitem{Song23Reaching}
Y.~Song, A.~Romero, M.~Mueller, V.~Koltun, and D.~Scaramuzza, ``Reaching the limit in autonomous racing: Optimal control versus reinforcement learning,'' \emph{Science Robotics}, p. adg1462, 2023.

\bibitem{kaufmann23champion}
E.~Kaufmann, L.~Bauersfeld, A.~Loquercio, M.~M{\"u}ller, V.~Koltun, and D.~Scaramuzza, ``Champion-level drone racing using deep reinforcement learning,'' \emph{Nature}, vol. 620, no. 7976, pp. 982--987, Aug 2023.

\bibitem{ferede2025one}
R.~Ferede, T.~Blaha, E.~Lucassen, C.~De~Wagter, and G.~C. de~Croon, ``One net to rule them all: Domain randomization in quadcopter racing across different platforms,'' \emph{arXiv preprint arXiv:2504.21586}, 2025.

\bibitem{song2021autonomous}
Y.~Song, M.~Steinweg, E.~Kaufmann, and D.~Scaramuzza, ``Autonomous drone racing with deep reinforcement learning,'' in \emph{2021 IEEE/RSJ International Conference on Intelligent Robots and Systems (IROS)}.\hskip 1em plus 0.5em minus 0.4em\relax IEEE, 2021, pp. 1205--1212.

\bibitem{kang2024autonomous}
Y.~Kang, J.~Di, M.~Li, Y.~Zhao, and Y.~Wang, ``Autonomous multi-drone racing method based on deep reinforcement learning,'' \emph{Science China Information Sciences}, vol.~67, no.~8, p. 180203, 2024.

\bibitem{romero2021model}
A.~Romero, S.~Sun, P.~Foehn, and D.~Scaramuzza, ``Model predictive contouring control for time-optimal quadrotor flight,'' \emph{IEEE Transactions on Robotics}, pp. 1--17, 2022.

\bibitem{romero2022replanningRAL}
A.~Romero, R.~Penicka, and D.~Scaramuzza, ``Time-optimal online replanning for agile quadrotor flight,'' \emph{IEEE Robotics and Automation Letters}, vol.~7, no.~3, pp. 7730--7737, 2022.

\bibitem{foehn2021time}
P.~Foehn, A.~Romero, and D.~Scaramuzza, ``Time-optimal planning for quadrotor waypoint flight,'' \emph{Science Robotics}, vol.~6, no.~56, p. eabh1221, 2021.

\bibitem{spica2018game}
R.~Spica, D.~Falanga, E.~Cristofalo, E.~Montijano, D.~Scaramuzza, and M.~Schwager, ``A game theoretic approach to autonomous two-player drone racing,'' \emph{arXiv preprint arXiv:1801.02302}, 2018.

\bibitem{pustilnik2025non}
M.~Pustilnik, A.~Loquercio, and F.~Borrelli, ``Non-normalized solutions of generalized nash equilibrium in autonomous racing,'' \emph{arXiv preprint arXiv:2503.12002}, 2025.

\bibitem{zhu2024sequential}
E.~L. Zhu and F.~Borrelli, ``A sequential quadratic programming approach to the solution of open-loop generalized nash equilibria for autonomous racing,'' \emph{arXiv preprint arXiv:2404.00186}, 2024.

\bibitem{lee2025champion}
H.~Lee, T.~Seno, J.~J. Tai, K.~Subramanian, K.~Kawamoto, P.~Stone, and P.~R. Wurman, ``A champion-level vision-based reinforcement learning agent for competitive racing in gran turismo 7,'' \emph{IEEE Robotics and Automation Letters}, 2025.

\bibitem{sims94creature}
K.~Sims, ``Evolving virtual creatures,'' in \emph{Proceedings of the 21st Annual Conference on Computer Graphics and Interactive Techniques}, ser. SIGGRAPH '94, 1994, p. 15–22.

\bibitem{jaderberg2019human}
M.~Jaderberg, W.~M. Czarnecki, I.~Dunning, L.~Marris, G.~Lever, A.~G. Castaneda, C.~Beattie, N.~C. Rabinowitz, A.~S. Morcos, A.~Ruderman \emph{et~al.}, ``Human-level performance in 3d multiplayer games with population-based reinforcement learning,'' \emph{Science}, vol. 364, no. 6443, pp. 859--865, 2019.

\bibitem{liu2019emergent}
S.~Liu, G.~Lever, J.~Merel, S.~Tunyasuvunakool, N.~Heess, and T.~Graepel, ``Emergent coordination through competition,'' \emph{arXiv preprint arXiv:1902.07151}, 2019.

\bibitem{cusumanorobust}
M.~F. Cusumano-Towner, D.~Hafner, A.~Hertzberg, B.~Huval, A.~Petrenko, E.~Vinitsky, E.~Wijmans, T.~W. Killian, S.~Bowers, O.~Sener \emph{et~al.}, ``Robust autonomy emerges from self-play,'' in \emph{Forty-second International Conference on Machine Learning}, 2025.

\bibitem{vinitsky2023learning}
E.~Vinitsky, R.~K{\"o}ster, J.~P. Agapiou, E.~A. Du{\'e}{\~n}ez-Guzm{\'a}n, A.~S. Vezhnevets, and J.~Z. Leibo, ``A learning agent that acquires social norms from public sanctions in decentralized multi-agent settings,'' \emph{Collective Intelligence}, vol.~2, no.~2, p. 26339137231162025, 2023.

\bibitem{ferede2023end}
R.~Ferede, C.~De~Wagter, D.~Izzo, and G.~C. de~Croon, ``End-to-end reinforcement learning for time-optimal quadcopter flight,'' \emph{arXiv preprint arXiv:2311.16948}, 2023.

\bibitem{ferede2024end}
R.~Ferede, G.~de~Croon, C.~De~Wagter, and D.~Izzo, ``End-to-end neural network based optimal quadcopter control,'' \emph{Robotics and Autonomous Systems}, vol. 172, p. 104588, 2024.

\bibitem{zhang2025learning}
D.~Zhang, A.~Loquercio, J.~Tang, T.-H. Wang, J.~Malik, and M.~W. Mueller, ``A learning-based quadcopter controller with extreme adaptation,'' \emph{IEEE Transactions on Robotics}, 2025.

\bibitem{zhao2024gate}
F.~Zhao, J.~Mei, J.~Zhou, Y.~Chen, J.~Chen, and S.~Li, ``Gate-aware online planning for two-player autonomous drone racing,'' \emph{arXiv preprint arXiv:2402.18021}, 2024.

\bibitem{spica2020real}
R.~Spica, E.~Cristofalo, Z.~Wang, E.~Montijano, and M.~Schwager, ``A real-time game theoretic planner for autonomous two-player drone racing,'' \emph{IEEE Transactions on Robotics}, vol.~36, no.~5, pp. 1389--1403, 2020.

\bibitem{di2023cooperative}
J.~Di, S.~Chen, P.~Li, X.~Wang, H.~Ji, and Y.~Kang, ``A cooperative-competitive strategy for autonomous multidrone racing,'' \emph{IEEE Transactions on Industrial Electronics}, vol.~71, no.~7, pp. 7488--7497, 2023.

\bibitem{wang2024dashing}
X.~Wang, J.~Zhou, Y.~Feng, J.~Mei, J.~Chen, and S.~Li, ``Dashing for the golden snitch: Multi-drone time-optimal motion planning with multi-agent reinforcement learning,'' \emph{arXiv preprint arXiv:2409.16720}, 2024.

\bibitem{vinyals2019grandmaster}
O.~Vinyals, I.~Babuschkin, W.~M. Czarnecki, M.~Mathieu, A.~Dudzik, J.~Chung, D.~H. Choi, R.~Powell, T.~Ewalds, P.~Georgiev \emph{et~al.}, ``Grandmaster level in starcraft ii using multi-agent reinforcement learning,'' \emph{nature}, vol. 575, no. 7782, pp. 350--354, 2019.

\bibitem{baker2019emergent}
B.~Baker, I.~Kanitscheider, T.~Markov, Y.~Wu, G.~Powell, B.~McGrew, and I.~Mordatch, ``Emergent tool use from multi-agent interaction,'' 2019.

\bibitem{yu2022surprising}
C.~Yu, A.~Velu, E.~Vinitsky, J.~Gao, Y.~Wang, A.~Bayen, and Y.~Wu, ``The surprising effectiveness of ppo in cooperative multi-agent games,'' \emph{NeurIPS}, vol.~35, pp. 24\,611--24\,624, 2022.

\bibitem{schulman2017proximal}
J.~Schulman, F.~Wolski, P.~Dhariwal, A.~Radford, and O.~Klimov, ``Proximal policy optimization algorithms,'' \emph{arXiv preprint arXiv:1707.06347}, 2017.

\bibitem{kunapuli2025leveling}
P.~Kunapuli, J.~Welde, D.~Jayaraman, and V.~Kumar, ``Leveling the playing field: Carefully comparing classical and learned controllers for quadrotor trajectory tracking,'' 2025.

\bibitem{forster2015system}
J.~F{\"o}rster, ``System identification of the crazyflie 2.0 nano quadrocopter,'' {B.S.} thesis, ETH Zurich, 2015.

\bibitem{mittal2023orbit}
M.~Mittal, C.~Yu, Q.~Yu, J.~Liu, N.~Rudin, D.~Hoeller, J.~L. Yuan, R.~Singh, Y.~Guo, H.~Mazhar, A.~Mandlekar, B.~Babich, G.~State, M.~Hutter, and A.~Garg, ``Orbit: A unified simulation framework for interactive robot learning environments,'' \emph{IEEE Robotics and Automation Letters}, vol.~8, no.~6, pp. 3740--3747, 2023.

\bibitem{khirodkar2018adversarial}
R.~Khirodkar and K.~M. Kitani, ``Adversarial domain randomization,'' \emph{arXiv preprint arXiv:1812.00491}, 2018.

\end{thebibliography}
}

\end{document}